\documentclass[conference]{IEEEtran}
\IEEEoverridecommandlockouts
\usepackage{cite}
\usepackage{amsmath,amssymb,amsfonts}
\usepackage{algorithmic}
\usepackage{graphicx}
\usepackage{textcomp}
\usepackage{xcolor}
\def\BibTeX{{\rm B\kern-.05em{\sc i\kern-.025em b}\kern-.08em
    T\kern-.1667em\lower.7ex\hbox{E}\kern-.125emX}}
\begin{document}

\title{A Time-Frequency Dual-Domain Multi-Scale Convolutional Neural Network for Bearing Fault Diagnosis under Strong Noise}

\author{\IEEEauthorblockN{1\textsuperscript{st} Yanxi Ding$^{\ast}$}
\IEEEauthorblockA{\textit{School of Engineering} \\
\textit{China University of Petroleum-Beijing at}\\
Karamay Campus \\
Karamay, China \\
2023015999@st.cupk.edu.cn}
\and
\IEEEauthorblockN{2\textsuperscript{nd} Tingyue Jia}
\IEEEauthorblockA{\textit{School of Engineering} \\
\textit{China University of Petroleum-Beijing at}\\
Karamay Campus \\
Karamay, China \\
2023016161@st.cupk.edu.cn}
}

\maketitle

\begin{abstract}
To address the degradation of bearing fault diagnosis accuracy under strong noise, this paper proposes a time-frequency dual-domain multi-scale convolutional neural network. The time-domain branch employs three parallel convolutional kernels to capture multi-scale impulse features, while the frequency-domain branch applies the Fast Fourier Transform to extract noise-robust spectral structure information. Features from both branches are fused for fault classification, yielding a compact model of 110,122 parameters. Experiments on the CWRU bearing dataset across seven signal-to-noise ratio levels demonstrate that the proposed method achieves 99.75\% accuracy under clean conditions and maintains 92.50\% at -4 dB SNR, representing a 7.25 percentage-point improvement over the single-domain baseline with monotonically increasing gains under stronger noise. Ablation experiments validate the independent performance contributions of the time-domain multi-scale branch and the frequency-domain branch. Comparative experiments against WDCNN, DRSN-CW, MCNN, and 1D-LeNet confirm the superiority of the proposed method under strong noise conditions.
\end{abstract}


\section{Introduction}

Rolling element bearings are among the most widely used critical components in rotating machinery, and their operating condition directly affects overall system safety and reliability. Statistical studies indicate that bearing-related faults account for approximately 45\% to 55\% of all rotating machinery failures \cite{1}, making bearing fault diagnosis a task of significant engineering importance. Vibration-based fault diagnosis has evolved from classical signal processing techniques to the deep learning paradigm \cite{2}, with convolutional neural networks (CNNs) emerging as the dominant approach due to their end-to-end feature learning capability, and the CWRU bearing dataset has served as a widely adopted public benchmark for evaluating these methods \cite{3}. In practical industrial environments, however, vibration signals are inevitably contaminated by strong background noise, and maintaining diagnostic accuracy under such conditions remains an open challenge.

Existing deep CNN-based diagnostic methods primarily extract fault features from time-domain waveforms. Under strong additive white Gaussian noise (AWGN), the impulse morphology of time-domain signals is directly corrupted, leading to a sharp decline in feature separability. Frequency-domain analysis offers a physically well-motivated complementary path: the power spectral density of AWGN is theoretically constant, merely raising the spectral baseline uniformly without introducing spurious local peaks; the structured spectral components induced by faults---including resonance bands, characteristic frequency harmonics, and modulation sidebands---retain their integrity even when the corresponding time-domain waveforms are submerged in noise. Building upon this physical insight, this paper proposes a time-frequency dual-domain multi-scale convolutional neural network: the time-domain branch employs three parallel convolutional kernels for multi-scale impulse feature extraction, while the frequency-domain branch applies the FFT followed by a lightweight convolutional subnetwork to extract spectral features; the two branches are fused in parallel for complementary dual-domain fault diagnosis.

This paper proposes a time-frequency dual-domain CNN architecture integrating parallel time-domain multi-scale convolution with frequency-domain FFT-based feature extraction. The time-domain branch employs three parallel convolutional kernels ($k$ = 3, 7, 15) to capture multi-scale impulse features, while the frequency-domain branch applies the FFT followed by a lightweight convolutional subnetwork to extract spectral structure information; features from both branches are concatenated and jointly trained through fully connected layers. The frequency branch accounts for only 2.5\% of the total parameters, achieving dual-domain information fusion at minimal architectural overhead. A systematic experimental evaluation is conducted on the CWRU bearing dataset across seven SNR levels, encompassing multi-SNR robustness analysis, comparison with four representative methods, ablation studies, and error analysis, with t-SNE feature clustering metrics serving as supplementary reference. Experimental results demonstrate that the proposed method consistently outperforms the single-domain baseline across all SNR levels and achieves 92.50\% accuracy under -4 dB strong noise, surpassing all compared methods and validating the effectiveness of the dual-domain design.

\section{Related Work}

Deep CNNs have achieved substantial progress in bearing fault diagnosis. WDCNN \cite{4} employs a wide first-layer convolutional kernel ($k$ = 64, stride 8) to enlarge the time-domain receptive field for capturing long-period impulse patterns. DRSN-CW \cite{5} embeds channel-wise learnable soft-thresholding functions within residual units, automatically estimating denoising thresholds through an attention mechanism for adaptive noise reduction in the time domain. MCNN \cite{6} adopts three parallel convolutional kernels ($k$ = 3, 5, 7) to simultaneously extract time-domain impulse features at different granularities, validating the effectiveness of multi-scale design for fault diagnosis. A compact 1D-CNN \cite{7} directly processes raw vibration signals through a simple convolutional-pooling-FC chain for end-to-end classification. A common characteristic of these methods is that feature extraction is performed entirely within the time domain: under strong AWGN, the impulse morphology of time-domain waveforms is directly corrupted. Regardless of whether wide kernels, adaptive thresholding, or multi-scale parallelism are employed, the fundamental loss of time-domain information cannot be remedied. This limitation constitutes the direct motivation for introducing a parallel frequency-domain branch in this work.

Multi-scale feature extraction is an important means of enhancing CNN representational capacity. In computer vision, the Inception architecture proposed by Szegedy et al. \cite{8} demonstrated the effectiveness of parallel multi-scale convolutions in capturing features at different receptive field sizes. This idea was subsequently transferred to one-dimensional vibration signals by methods such as MCNN \cite{6}, enabling multi-granularity time-domain analysis through three differently sized convolutional kernels. Frequency-domain features have a long history of application in rotating machinery fault diagnosis---FFT magnitude spectra, envelope spectra, and spectral kurtosis are core tools in classical diagnostic approaches \cite{1}. However, traditional frequency-domain methods typically rely on manual feature engineering and expert prior knowledge, and are often employed as preprocessing steps independent of the learning model. The key distinctions between this work and prior approaches lie in three aspects. First, the frequency-domain branch and the time-domain branch receive the same input tensor during the forward pass, compute in parallel, and are fused through concatenation --- the FFT resides inside the network rather than serving as an external preprocessing stage. Second, the convolutional kernels of the frequency branch, as learnable parameters, automatically converge to spectral templates corresponding to fault characteristic frequencies through backpropagation, replacing manual feature selection under fixed mathematical transforms such as envelope spectra and spectral kurtosis. Third, the concatenated dual-domain features are jointly supervised by the classification loss, with gradients backpropagating through both branches simultaneously, thereby enabling the frequency branch to adaptively learn spectral patterns that complement the representations already captured by the time-domain branch.

\section{Method}

The input of the proposed time-frequency dual-domain multi-scale convolutional neural network in this paper is a single-channel raw vibration signal with a sampling frequency of 48 kHz and a window length of 1,024 points (approximately 21.3 ms). After Z-score normalization, it is sent in parallel to the time-domain multi-scale branch and the frequency-domain branch. The time-domain branch outputs a 128-dimensional feature vector, while the frequency-domain branch outputs a 32-dimensional spectral feature vector. These two are concatenated along the feature dimensions to obtain a 160-dimensional joint feature, which is then mapped to 10 types of fault probabilities through two layers of fully connected networks. All convolutional layers are equipped with batch normalization and ReLU activation functions, and downsampling uses maximum pooling. The total number of parameters of the model is 110,122, among which the frequency-domain branch only accounts for 2,768 parameters, achieving dual-domain information fusion with minimal architectural overhead.

\subsection{Time-Domain Multi-Scale Branch}

The time-domain branch replaces the conventional single-kernel design with three parallel first-layer convolutional kernels. Given a normalized input signal, the three kernels perform 1-D convolution in parallel:
\begin{equation}
	z_k=ReLU(Convld(x;C_{in}=1,C_{out}=16,kernel\_size=k))
\end{equation}

The three output feature maps are concatenated along the channel dimension, forming a 48-channel tensor. The three kernel sizes correspond to distinct physical time scales: $k$ = 3 ($\approx$0.063 ms) captures the transient rising and falling edges of impulses; $k$ = 7 ($\approx$0.146 ms) matches the complete envelope of a single impulse event; and $k$ = 15 ($\approx$0.313 ms) spans cross-period modulation patterns. This multi-scale design adds only 256 parameters (an approximately 0.3\% increase over the single-kernel baseline of 97,434 parameters), simultaneously extracting impulse features at three time scales at negligible capacity cost.

The concatenated feature map is subsequently processed by two convolutional blocks. Each block consists of two Conv1d($k$ = 3) layers with batch normalization and ReLU activation. A max-pooling layer with stride 4 performs downsampling between the blocks. Global average pooling compresses the feature map into a 128-dimensional time-domain feature vector, which encodes the impulse morphology of the input signal across three physical time scales.

\subsection{Frequency-Domain Branch}

The design of the frequency-domain branch is motivated by the spectral characteristics of additive white Gaussian noise: the power spectral density of AWGN is theoretically constant, merely raising the spectral baseline uniformly. In contrast, fault-induced resonance bands, characteristic frequency harmonics, and modulation sidebands manifest as structured local energy concentrations in the spectrum, and these spectral structures remain intact even when the corresponding time-domain waveforms are submerged in noise. The frequency-domain branch is designed to automatically capture such structured fault patterns from the spectrum through learnable convolutional kernels.

First, the real-valued Fast Fourier Transform (RFFT) is applied to the normalized input signal, and the first 512 frequency bins of the positive half-spectrum are retained:
\begin{equation}
	X=|RFFT(x)|_{[0:512]}\in R^{512}
\end{equation}

The spectral vector is processed by two 1-D convolutional layers. The first layer, Conv1d, maps the single-channel spectrum to 16 feature channels; after batch normalization and ReLU activation, max pooling with stride 4 performs downsampling. The second layer, Conv1d, further expands the representation to 32 channels. Following the same normalization–activation–pooling sequence, global average pooling compresses the output into a 32-dimensional frequency-domain feature vector. The convolutional kernels of the frequency branch adaptively learn spectral templates corresponding to fault characteristic frequencies through backpropagation, producing high activation responses to matched spectral patterns while suppressing noise-dominated flat spectral regions---a mechanism that can be understood as data-driven implicit matched filtering.

\subsection{Feature Fusion and Classification}

The time-domain feature and the frequency-domain feature are concatenated along the feature dimension to form a 160-dimensional joint feature vector:
\begin{equation}
	f=\begin{bmatrix}
		f_t \\
		f_f
	\end{bmatrix} \in R^{160}
\end{equation}

The joint feature is mapped to fault categories through two fully connected layers: the first layer, FC, projects the dual-domain features into a 64-dimensional latent space with ReLU activation for nonlinearity; the second layer, FC, outputs the unnormalized logits for the 10 fault classes, which are converted to class probability distributions via the Softmax function. The concatenation operation preserves the independence of the two feature spaces---time-domain channels encode waveform morphology while frequency-domain channels encode spectral energy distribution, two physically distinct types of information. The fully connected layers adaptively learn combination weights for the cross-domain features during training, eliminating the need for manually specified fusion coefficients.

\section{Experiments and Results}

\subsection{Dataset and Experimental Setup}

Experiments are conducted on the publicly available Case Western Reserve University (CWRU) bearing dataset. Vibration signals acquired from the drive-end accelerometer at a sampling rate of 48 kHz under a 1 HP load (approximately 1,772 r/min) are used, covering 10 bearing conditions: one normal state and nine fault modes spanning three fault types (Ball, Inner Race, Outer Race) at three damage diameters (0.007, 0.014, and 0.021 inches). Signals are segmented using a non-overlapping sliding window of 1,024 points (approximately 21.3 ms) with Z-score normalization, yielding 4,745 samples in total. A stratified 75/25 train/test split is applied with a fixed random seed of 42, resulting in 3,558 training and 1,187 test samples. The class distribution is shown in TABLE I.

\begin{table}[!ht]
	\centering
	\caption{Composition of the CWRU bearing dataset}
	\begin{tabular}{cccccc}
		\hline
		Class	&	Fault Type	&	Diameter/mil	&	Train	&	Test	&	Total \\
		\hline
		Ball\_007	&	Ball	&	7	&	356	&	119	&	475 \\
		Ball\_014	&	Ball	&	14	&	355	&	119	&	474 \\
		Ball\_021	&	Ball	&	21	&	356	&	119	&	475 \\
		IR\_007	&	Inner Race	&	7	&	355	&	119	&	474 \\
		IR\_014	&	Inner Race	&	14	&	358	&	119	&	477 \\
		IR\_021	&	Inner Race	&	21	&	355	&	118	&	473 \\
		Normal	&	Normal	&	--	&	354	&	118	&	472 \\
		OR\_007	&	Outer Race	&	7	&	356	&	119	&	475 \\
		OR\_014	&	Outer Race	&	14	&	355	&	118	&	473 \\
		OR\_021	&	Outer Race	&	21	&	358	&	119	&	477 \\
		Total	&	--	&	--	&	3,558	&	1,187	&	4,745 \\
		\hline
	\end{tabular}
\end{table}

Noise protocol: Additive white Gaussian noise is added exclusively to the test set at seven SNR levels: Clean (no noise), 6 dB, 4 dB, 2 dB, 0 dB, -2 dB, and -4 dB. At -4 dB, the noise power is approximately 2.51 times the signal power, simulating extreme noise conditions. Deterministic noise seeds ensure identical test conditions across all models. Training configuration: Adam optimizer (learning rate 0.001), batch size 64, 30 epochs, global random seed 42. Evaluation metrics are test accuracy and macro-averaged F1 score. All experiments are conducted on CPU. Online random noise augmentation is applied to the training set during training to improve generalization to the noise distribution.

\subsection{Multi-SNR Robustness}

To systematically evaluate the performance of the dual-domain architecture under different noise intensities, three progressive model variants are defined: M1 (single-kernel baseline, $k$=3, 97,434 parameters), M2 (M1 + L1 multi-scale kernels $k$=3/7/15, 97,690 parameters), and M3 (M2 + frequency branch, i.e., the proposed full model, 110,122 parameters). Each model is independently trained and tested at each SNR level, totaling 21 runs.

\begin{figure}[!ht]
	\centering
	\includegraphics[width=\linewidth]{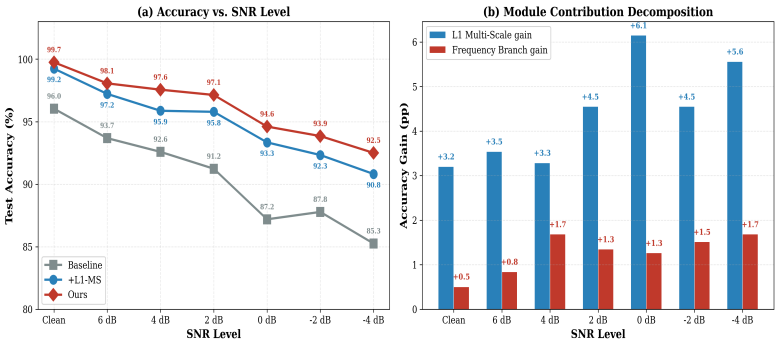}
	\caption{Multi-SNR robustness analysis.}
	\label{fig:1}
\end{figure}

From Fig. \ref{fig:1}, the following findings can be obtained. The dual-domain model in this paper outperforms the single-domain baseline at all SNR levels, and the advantage increases with the enhancement of noise. M3 achieves an accuracy rate of 99.75\% under the Clean condition, dropping to 92.50\% at -4 dB, with a total degradation of only 7.25 percentage points; meanwhile, M1 decreases from 96.04\% to 85.26\%, with a degradation of 10.78 percentage points. The gain of M3 relative to M1 increases from +3.71 percentage points (Clean) to +7.25 percentage points (-4 dB), and the increasing trend of the gain eliminates the alternative explanation of increased parameter quantity - if it is only a capacity effect, the gain should be approximately constant across all SNR levels. The contribution of the frequency domain branch increases with the enhancement of noise: under the Clean condition, the frequency domain gain is only +0.51 percentage points, and at this time, the time domain features themselves already have sufficient discriminative power; as the SNR decreases to -4 dB, it increases to +1.68 percentage points, which is consistent with the physical characteristics of AWGN - the stronger the noise, the more prominent the advantage of the frequency domain over the time domain structure. The multi-scale time-domain branch provides a stable base gain of 3 to 6 percentage points across all SNR levels, with a fluctuation amplitude of approximately 2.95 percentage points across SNR, indicating that multi-scale features capture is the basic component of noise robustness, while the incremental contribution of the frequency domain features above this platform increases with the deterioration of noise, and the two form an effective noise-resistant synergy.

\subsection{Comparison with Existing Methods}

\begin{figure}[!ht]
	\centering
	\includegraphics[width=\linewidth]{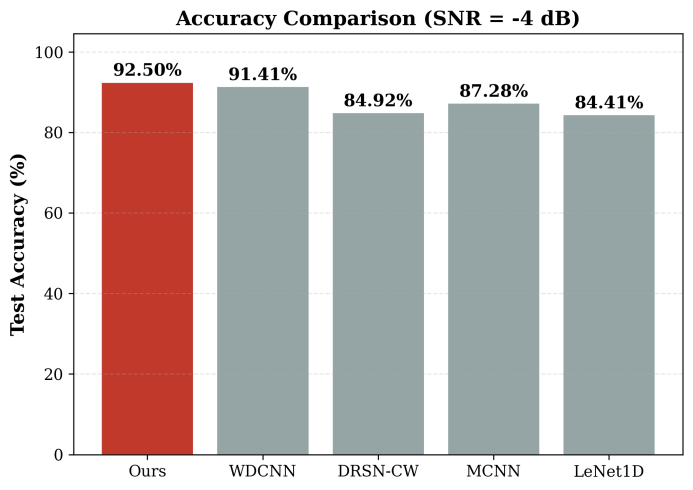}
	\caption{Comparison with existing methods.}
	\label{fig:2}
\end{figure}

At SNR = -4 dB, the proposed method is compared against four published methods: WDCNN, DRSN-CW, MCNN, and 1D-LeNet. All comparison methods are rigorously reproduced under identical data splits, noise seeds, and training protocols to ensure fairness. Results are presented in Fig. \ref{fig:2}.

The proposed method achieves the highest accuracy of 92.50\%, surpassing the second-best WDCNN by 1.09 pp and MCNN by 5.22 pp. Notably, MCNN also adopts parallel multi-scale convolution ($k$=3/5/7) but without a frequency branch; its accuracy of 87.28\% is comparable to our M2 variant (90.82\%). The 5.22 pp gap between the proposed method and MCNN is almost entirely attributable to the introduction of the frequency branch, independently validating the effectiveness of the dual-domain design from an external benchmark. DRSN-CW achieves only 84.92\% despite its adaptive soft-thresholding mechanism for time-domain denoising, suggesting that reliably estimating thresholds from waveforms under extreme noise is inherently challenging---a finding that supports our design philosophy of explicit frequency-domain transformation over implicit time-domain denoising. 1D-LeNet achieves 84.41\% with merely 11,462 parameters, demonstrating the competitiveness of lightweight architectures, yet lags behind the proposed method by 8.09 pp, highlighting the limitations of pure depth stacking under strong noise.

\subsection{Ablation Study}

\begin{figure}[!ht]
	\centering
	\includegraphics[width=\linewidth]{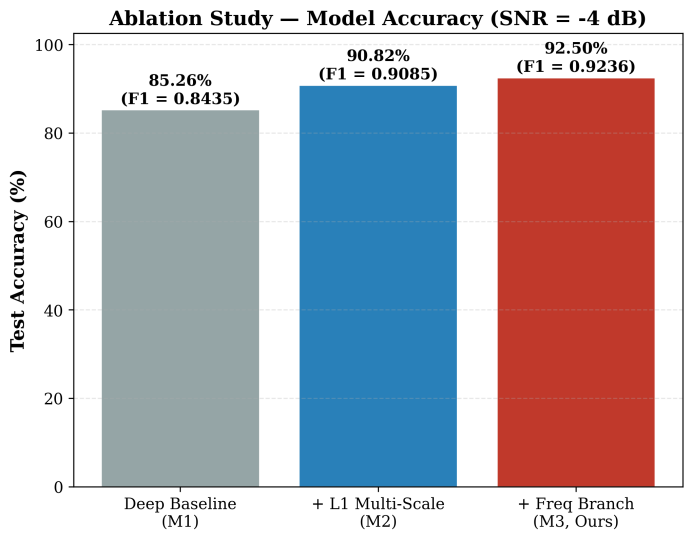}
	\caption{Ablation analysis.}
	\label{fig:3}
\end{figure}

An ablation study is conducted at SNR = -4 dB by incrementally adding modules following the M1$\rightarrow$M2$\rightarrow$M3 progression to quantify the independent contribution of each component. Results are shown in Fig. \ref{fig:3}. The ablation results demonstrate that the time-domain multi-scale branch contributes a gain of +5.56 pp, and the frequency-domain branch contributes an additional +1.68 pp, together accounting for the total M1→M3 improvement of 7.25 pp. Both branches make independent positive contributions to model performance, validating the effectiveness of each component.In terms of parameter efficiency, the multi-scale parallel kernels add only 256 parameters (approximately 0.3\% increase) for a +5.56 pp gain, and the frequency convolutional layers contribute only 2,768 parameters (2.5\% of the total) for a +1.68 pp gain, embodying a design principle of architectural innovation rather than parameter scaling.

\subsection{Feature Space Analysis}

At SNR = -4 dB, feature vectors were extracted from the penultimate layer of M1 (128-d GAP) and M3 (160-d concatenation layer) and projected to a 2-D plane using t-SNE (perplexity 50, PCA initialization, 2,000 iterations). Cluster quality was quantified by the Silhouette Score \cite{9} and the Davies-Bouldin Index (DBI). M1 achieved a Silhouette Score of 0.2415 and a DBI of 1.9050; M3 achieved a Silhouette Score of 0.2866 and a DBI of 1.5323.

\subsection{Error Analysis}

\begin{figure}[!ht]
	\centering
	\includegraphics[width=\linewidth]{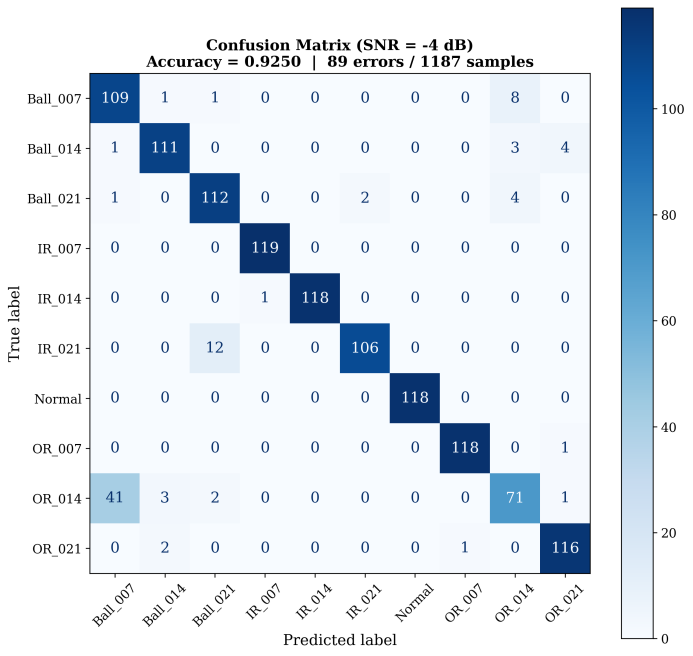}
	\caption{Confusion matrix.}
	\label{fig:4}
\end{figure}

The proposed method produces 89 misclassifications out of 1,187 test samples (7.5\%) at SNR = -4 dB. The confusion matrix and per-class accuracy are shown in Fig. \ref{fig:4}. The matrix exhibits a strong diagonal structure, with four classes achieving nearly perfect accuracy: IR\_007 (100\%), Normal (100\%), IR\_014 (99.16\%), and OR\_007 (99.16\%). Errors are concentrated in a few physically interpretable class pairs: OR\_014 $\rightarrow$ Ball\_007 (41 samples, 46.1\% of all errors) constitutes the most prominent confusion pair. Its physical origin lies in the relatively weak modulation signature of outer-race faults at the 6 o'clock load zone position, coupled with the limited frequency resolution of 46.875 Hz under the 21.3 ms short window, creating a risk of spectral aliasing between low-order harmonics of the outer-race fault characteristic frequency (BPFO, $\approx$106 Hz at 1 HP) and the ball spin frequency (BSF, $\approx$70 Hz). IR\_021 $\rightarrow$ Ball\_021 (12 samples, 13.5\%) represents confusion between faults of the same severity (0.021 inches) at different locations --- the broadband impulse energy generated by large-size faults may obscure location-specific information within the short time window. Ball\_007 $\rightarrow$ OR\_014 (8 samples, 9.0\%) forms an asymmetric bidirectional confusion pattern with the aforementioned OR\_014 $\rightarrow$ Ball\_007 pair.

The above error patterns possess clear physical interpretability, indicating non-random misclassifications rather than noise-induced random errors. They suggest the following directions for improvement: extending the analysis window to increase frequency resolution (to mitigate harmonic aliasing); incorporating multi-sensor fusion to disambiguate same-severity, different-location faults; and implementing a Softmax confidence-based rejection threshold for the OR\_014–Ball\_007 pair, deferring high-uncertainty samples for manual review.

\section{Conclusion}

This paper proposes a time-frequency dual-domain multi-scale convolutional neural network to address the core challenge of severe diagnostic accuracy degradation under strong noise in bearing fault diagnosis. At the methodological level, the time-domain branch employs three parallel convolutional kernels ($k$ = 3, 7, 15) that respectively capture the transient edges of impulses, the complete envelope of a single impulse event, and cross-period modulation patterns --- each kernel size corresponding to a physically meaningful time scale. The frequency-domain branch first maps the signal to the frequency domain via the real-valued Fast Fourier Transform (RFFT), then applies a lightweight 1-D convolutional subnetwork to automatically learn spectral templates corresponding to fault characteristic frequencies. Its working mechanism can be understood as data-driven implicit matched filtering: producing high activation responses to structured spectral components while suppressing the flat spectral baseline introduced by AWGN. Features from the two branches are concatenated along the feature dimension and adaptively fused through fully connected layers, eliminating the need for manually specified cross-domain weighting. The frequency branch accounts for only 2,768 parameters (2.5\% of the total 110,122), obtaining complementary frequency-domain information at minimal architectural overhead --- embodying a design philosophy of architectural innovation rather than parameter scaling.

At the experimental level, a five-dimensional systematic evaluation was conducted on the CWRU bearing dataset across seven SNR levels. The proposed dual-domain model consistently outperforms the single-domain baseline from Clean to -4 dB, with the performance gain increasing monotonically with noise intensity from +3.71 pp to +7.25 pp, ruling out a mere capacity effect as the alternative explanation. At -4 dB extreme noise, the proposed method achieves 92.50\% accuracy, significantly surpassing four representative methods: WDCNN (91.41\%), MCNN (87.28\%), DRSN-CW (84.92\%), and 1D-LeNet (84.41\%). The ablation study demonstrates that the time-domain multi-scale branch contributes a gain of +5.56 pp and the frequency-domain branch contributes an additional +1.68 pp, together accounting for the total improvement of 7.25 pp, thereby validating the independent contribution of each component. The t-SNE feature clustering analysis yields a Silhouette Score improvement from 0.2415 to 0.2866 and a Davies-Bouldin Index reduction from 1.9050 to 1.5323. Error analysis reveals that the 7.5\% misclassification rate is concentrated in a few physically interpretable class pairs, with the OR\_014–Ball\_007 confusion accounting for 46.1\% of all errors, representing non-random misclassifications that point to concrete improvement directions including spectral aliasing mitigation and enhanced frequency resolution under short analysis windows.

The present work has systematically validated the proposed method on the CWRU dataset under a 1 HP load condition. Building upon these findings, further investigations may be pursued along the following directions.

First, extending the method to CWRU multi-load (0–3 HP) and variable-speed conditions, as well as to the Paderborn University bearing dataset \cite{10} and the XJTU-SY run-to-failure dataset \cite{11}, to systematically characterize cross-domain generalization under domain shift. Second, conducting multi-seed repeated experiments to establish statistical confidence intervals and enable significance testing of inter-model performance differences, thereby strengthening the statistical reliability of the conclusions. Third, investigating the optimal trade-off between segment length and frequency resolution, with zero-padding before the RFFT as a candidate technique to increase effective resolution and further mitigate low-order harmonic aliasing. Fourth, applying knowledge distillation with a compact student network to reduce inference latency for real-time edge deployment while preserving diagnostic accuracy. Finally, exploring the potential of complex-valued FFT that retains phase information for the frequency branch, as well as multi-sensor fusion of drive-end and fan-end accelerometers for disambiguating same-severity, different-location faults.

\end{document}